\documentclass{article}
\usepackage{ijcai21}
\usepackage{kotex}
\usepackage{times}
\usepackage{soul}
\usepackage{url}
\usepackage[draft]{hyperref}
\usepackage[utf8]{inputenc}
\usepackage[small]{caption}
\usepackage{graphicx}
\usepackage{amsmath}
\usepackage{amsthm}
\usepackage{booktabs}
\usepackage{algorithm}
\usepackage{algorithmic}
\usepackage{subfigure}
\usepackage{amssymb}
\usepackage{multirow}

\usepackage{array}

\usepackage[flushleft]{threeparttable}
\usepackage{xcolor}

\newcommand{\ie}{{\em i.e.}}           
\newcommand{\etc}{{\em etc.}}         

\newcommand{\bM}{{\bf M}}

\newcommand{\bX}{{\bf X}}

\begin{document}

\title{Multi-view Integration Learning for Irregularly-sampled Clinical Time Series}

\author{
Yurim Lee$^{1,*}$
\and
Eunji Jun$^{1,}$\footnote{Equally contributed}\And
Heung-Il Suk$^{1,2,}$\footnote{Corresponding author}
\affiliations
$^1$Department of Brain and Cognitive Engineering, Korea University\\
$^2$Department of Artificial Intelligence, Korea University\\
Seoul 02841, Republic of Korea
\emails
\{yurimalee, ejjun92, hisuk\}@korea.ac.kr
}


\maketitle

\begin{abstract}
Electronic health record (EHR) data is sparse and irregular as it is recorded at irregular time intervals, and different clinical variables are measured at each observation point. In this work, we propose a multi-view features integration learning from irregular multivariate time series data by self-attention mechanism in an imputation-free manner. Specifically, we devise a novel multi-integration attention module (MIAM) to extract complex information inherent in irregular time series data. In particular, we explicitly learn the relationships among the observed values, missing indicators, and time interval between the consecutive observations, simultaneously. The rationale behind our approach is the use of \emph{human knowledge} such as what to measure and when to measure in different situations, which are indirectly represented in the data. In addition, we build an attention-based decoder as a missing value imputer that helps empower the representation learning of the inter-relations among multi-view observations for the prediction task, which operates at the training phase only. We validated the effectiveness of our method over the public MIMIC-III and PhysioNet challenge 2012 datasets by comparing with and outperforming the state-of-the-art methods for in-hospital mortality prediction. 
\end{abstract}

\section{Introduction}
 Electronic health record (EHR) is a collection of patients' health data such as coded diagnoses, vital signs and lab test records, procedures, and textual narratives, \etc\ Over the last decade, there has been significant progress on developing deep learning models using multivariate EHR time series owing to its abundance of health big datasets. However, learning appropriate EHR representations is challenging for many classical machine learning models, due to the nature of EHR that is irregularly sampled, where observations are measured at different time points determined by the type of measurement, the patient's health status, and the availability of clinical staff. Different number of observations and a lack of temporal alignment across data invalidate the use of machine learning models that assume a fixed-dimensional feature space.

The classical machine learning approaches for handling the irregularly sampled time series are mostly based on convolutional neural network (CNN), recurrent neural network (RNN), and more recently attention-based methods, showing their superiority in healthcare target tasks. Specifically, CNN is usually applied to capture the local temporal characteristics of clinical data using the predefined kernel on a small chunk of EHR features for identifying local motifs, \ie, co-occurrences of diseases \cite{Deepr}, and patients similarity \cite{timefusionCNN}. But those works have limitations on modeling local temporal dependencies in the predefined kernel size, not global ones.

RNN-based methods have been the de-facto solution to deal with clinical time series data, as RNNs can manage various lengths of sequential data. But conventional RNN methods are designed to handle data with a constant time interval between consecutive time series, thus leading to suboptimal performance for irregular time interval. To address this challenge, the widely-used approach is to convert irregularly sampled time series data into regularly sampled time series, \ie, \emph{temporal discretization} \cite{lipton2016directly,MGPRNN} and feed this fixed-dimensional vector to RNNs. Nonetheless, it requires ad-hoc choices on the window size and the aggregation function that deals with values falling within the same window. Similar to discretization methods, interpolation methods \cite{MGPRNN,InterpNet1} require specifying discrete reference time points. Instead of using all available observations in the input to replace the interpolants at these time points, it may inevitably introduce additional noise or information loss due to assumption of a fixed time interval.

A better approach for handling irregular time series is to directly model the unequally spaced original data. Compared to the conventional RNN that relies on discrete time, ordinary differential equation (ODE)-based recurrence models \cite{odelstm} were proposed to handle non-uniform time intervals and remove the need to aggregate observations into equally spaced intervals by generalizing hidden state transitions in RNN to continuous time dynamics by ODEs. Another alternative is to exploit the \emph{source of missingness} such as missing indicators and time interval for modeling informative missingness pattern. The existing works \cite{GRUD,lipton2016directly,tlstm,attain,tan2020data} used either missing indicators or time interval, and applied the heuristic decaying function such as a monotonically non-increasing function, without learning representations for missingness.

More recently, attention-based methods \cite{retain,choi2017gram,ma2018kame,AttnDiag,horn2020set} have been used to deal with irregular sampling. In particular, self-attention models \cite{vaswani2017attention} offered computational advantages over RNNs due to their fully parallelized sequence processing. Several works based on self-attention mechanism have applied a simple modified self-attention such as a masked attention \cite{AttnDiag}, or replaced the positional encoding with time encoding and concatenated encoding vector and missing indicators \cite{horn2020set}.

To address the aforementioned challenges and limitations, this paper proposes a novel method to jointly learn deep representations of multi-view features from irregular multivariate time series data by using the self-attention mechanism in an imputation-free manner. Specifically, we devise a novel multi-view integration attention module (MIAM) to learn complex missing patterns by integrating missing indicators and time interval, and further combine observation and missing pattern in the representation space through a series of self-attention blocks. On top of the MIAM module, we build an attention-based decoder as a missing data imputer that helps empower the representation learning of the inter-relations among multi-view observations for the prediction task, which operates at the training phase only. As a result, model complexity is reduced, while removing the need to impute missing data at the same time. We showed that our proposed method outperformed state-of-the-art methods for in-hospital mortality prediction on real-world EHR datasets: MIMIC-III and PhysioNet 2012 challenge datasets.

\begin{figure*}
	\centering
        \includegraphics[width=\textwidth]{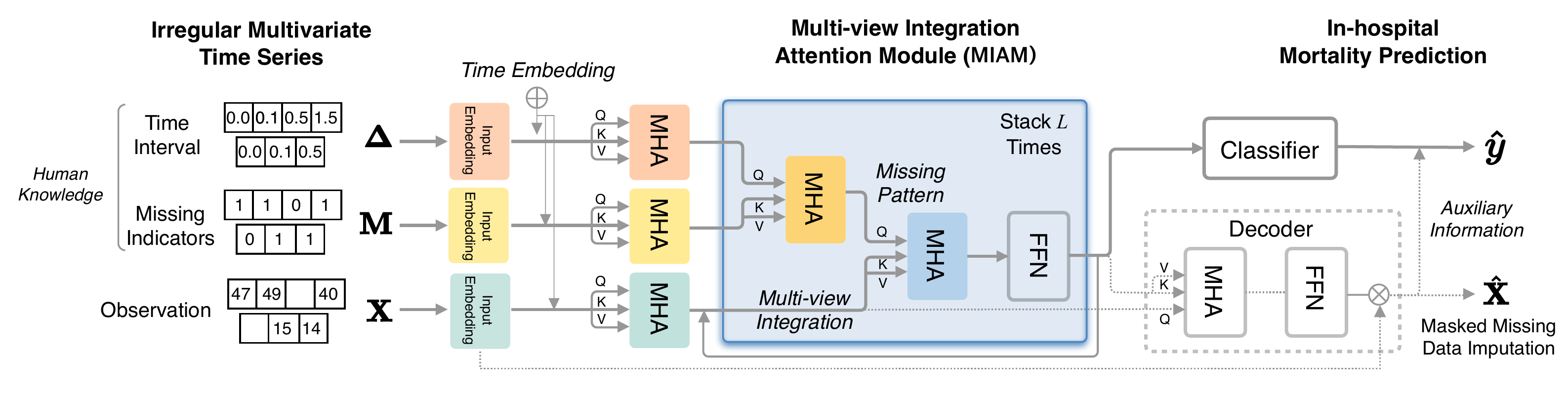}
        \caption{The overall framework of the proposed multi-view integration learning method for irregularly-sampled clinical time series. The solid lines represent a classification process, and dotted lines represent an auxiliary missing data imputation process. For input embedding and MHA, each observation and missing indicators have individual learnable weights, and two integration MHAs also have separate weights.}
        \label{fig:framework}
\end{figure*}

\section{Related Work}
\subsubsection{Irregular Time Series Modeling}
In order to accommodate irregularly sampled time series data, the widely-used approach is to discretize time into consecutive, non-overlapping uniform intervals, \ie, \emph{temporal discretization} \cite{lipton2016directly,MGPRNN}, which enables the use of models that operate on fixed-dimensional vectors. In the meantime, discretization reduces the problem from irregular time series data modeling into missing data imputation. The simple imputation approaches range from simple zero imputation and forward filling to more sophisticated deep learning approaches \cite{cao2018brits} including generative adversarial network (GAN) \cite{yoon2018gain} and variational autoencoder (VAE) \cite{fortuin2020gp,jun2019stochastic}, \etc\ However, this explicit imputation of missing data during discretization mostly depends on heuristic or unsupervised method that will not be universally applicable, and further needs to consider uncertainty \cite{9177349} for application of downstream clinical tasks.

Similar to discretization methods, interpolation methods require specifying discrete reference time points. A multi-task Gaussian process (MGP-RNN) model \cite{MGPRNN} conducted a probabilistic interpolation by transforming irregular time series into a more uniform representation on evenly spaced reference time points, and feeding the latent function values into RNNs. While it provides uncertainty, there are limitations on a predefined time interval between reference time points and limited expressiveness of the model from a sum of separable kernel function. Instead, the interpolation-prediction network (IPNet) \cite{InterpNet1} learned an optimal time interval for deterministic interpolation. The interpolation network first interpolated irregular time series for each variable, and then merged all time series across every variable. However, IPNet may unavoidably introduce additional noise or information loss because it was also expected to assume fixed time interval.

Accordingly, recent interest of irregularity-based methods is learning representations directly from multivariate sparse and irregularly sampled time series as input without the need for imputation \cite{zaheer2017deep,horn2020set}. To this end, ODE-LSTM model \cite{odelstm} was proposed to generalize hidden state transitions in RNN to continuous time dynamics by ODEs, compared to conventional RNN that depends on discrete time. In this way, it deals with continuous time observations within the LSTM network, enabling cells to handle non-uniform time intervals and remove the need to aggregate observations into equally spaced intervals.

\subsubsection{Missing Patterns Modeling}
For modeling informative missingness, \cite{GRUD} added a temporal decay derived from time interval to the input variables and hidden states, and directly included both observation and masking indicators inside the GRU architecture. Also, \cite{lipton2016directly} used hand-engineered features derived from the response indicator such as mean and standard deviation of missing indicators for each time series. \cite{tlstm} proposed a time-aware LSTM (T-LSTM) that decomposes the cell memory into short and long-term memories and decays the short-term memory by weights transformed from time intervals, while retaining the long-term memories. Specifically, a monotonically non-increasing function is heuristically chosen as a decaying function. Similarly, ATTAIN \cite{attain} decayed short-term memory by using both time intervals and weights generated from the attention mechanism by considering several previous events, not just one previous event. DATA-GRU \cite{tan2020data} also introduced a time-aware mechanism to a GRU for handling irregular time interval, and further devised a dual attention mechanism to deal with missing values in both data-quality and medical-knowledge views.

\subsubsection{Attention Mechanism in Irregular Time Series Modeling}
Several recent models have leveraged attention mechanisms as their fundamental approach to dealing with irregular sampling. For example, RETAIN \cite{retain} learned an interpretable representation of irregularly sampled time series by two-level RNNs, generating visit-level and variable-level attentions. To learn robust representations of EHR data, \cite{choi2017gram} introduced a graph-based attention method. Similarly, \cite{ma2018kame} proposed a knowledge-based attention mechanism to learn embeddings for nodes in the knowledge graph. Based on the self-attention \cite{vaswani2017attention}, \cite{AttnDiag} proposed a SAnD model that uses a masked self-attention specifying how far the attention model looks into the past and a dense interpolation for capturing temporal dependencies. In addition, \cite{horn2020set} replaced the positional encoding with time encoding and concatenated encoding vector and missing indicators.

\section{Method}
In this section, we present the proposed method for multi-view features integration learning of irregular multivariate EHR time series for in-hospital mortality prediction task. First, we introduce the notations for multivariate time series data, and then describe our proposed method that consists of (i) input and time embedding, (ii) multi-view integration learning, (iii) mortality prediction as a binary classification, and (iv) auxiliary imputation for masked missing data. The overall architecture is shown in Figure \ref{fig:framework}.

\subsection{Data Representation}
For each subject $n$, given a set of $D$-dimensional multivariate time series in $\mathbf{t}^{(n)}={[t_1,...,t_j,...,t_{T_n}}]$ time points of length $T_{ n}$, we denote an observation time series as $\mathbf{X}^{(n)} =[\mathbf{x}_{t_1}^{({ n})},...,\mathbf{x}_{t_j}^{({ n})},..., \mathbf{x}_{t_{T_{ n}}}^{({ n})}]^\top \in \mathbb{R}^{T_{ n}\times D}$, where $\mathbf{x}_{t_j}^{({ n})}\in\mathbb{R}^D$ represents the $t_j$-th observation of all variables, and $x_{t_j, d}^{({ n})}$ is the element of the $d$-th variable in $\mathbf{x}_{t_j}^{({ n})}$. In this setting, since the time series $\mathbf{X}^{({ n})}$ includes missing values, we introduce the masking vector across time series, ${\mathbf{M}^{({ n})}}=[\mathbf{m}_{t_1}^{({ n})},...,\mathbf{m}_{t_j}^{({ n})}, ..., \mathbf{m}_{t_{T_{ n}}}^{({ n})}]^\top\in\mathbb{R}^{T_{ n}\times D}$, which has the same size of $\mathbf{X}^{({ n})}$ to mark which variables are observed or missing. Specifically, we have $m_{t_j, d}^{({ n})}=1$ if $x_{t_j,d}^{({ n})}$ is observed, otherwise, $m_{t_j,d}^{({ n})}=0$. If the observation is missing, the input is set to zero for that dimension. For each variable $d$, we also present the \textit{time interval} $\mathbf{\Delta}^{({ n})}=[\boldsymbol{\delta}_{t_1}^{({ n})},...,\boldsymbol{\delta}_{t_j}^{({ n})}, ..., \boldsymbol{\delta}_{t_{T_{ n}}}^{({ n})}]^\top\in\mathbb{R}^{T_{ n}\times D}$,  { where} $\delta_{t_j, d}^{({ n})}\in \mathbb{R}$ is defined as:
\begin{equation}\label{eq1}
    \delta_{t_j, d}^{({ n})} =
    \begin{cases}
        t_j - t_{j-1} +  \delta_{t_{j-1}, d}^{({ n})}, & t_j>1, m_{t_{j-1}^{({ n})}, d}=0\\
        t_j - t_{j-1}, & t_j>1, m_{t_{j-1}^{({ n})}, d}=1\\
        0, & t_j = 1.\\
    \end{cases}
\end{equation}
In this paper, given a clinical time series dataset $\mathcal{D}=\{(\mathbf{X}^{({ n})}, \mathbf{M}^{({ n})}, \mathbf{\Delta}^{({ n})})\}_{n=1}^N$ for $N$ subjects, we { construct a mapping function to} predict the mortality labels $(y_{1}, \cdots, y_{N})$ in the binary classification problem.

For uncluttered, we will use functional notation that represents information about a particular patient, { omitting} superscript $(n)$ for $n$-th subject.

\subsection{Multi-view Integration Learning}
\label{subsec:ModelName}
{ The key feature of missing data is that there may be information conveyed by missingness itself, and ignoring this dependence may lead incorrect predictions. The existing works \cite{GRUD,lipton2016directly,tlstm,attain,tan2020data} leveraged these sources of missingness,  \ie, missing indicators and time interval, and applied the heuristic decaying function for their use without learning its representation. However, using inappropriate modeling for missingness may lead to unreliable assessment of feature importance and model that is not robust to measurement changes.

Motivated by this observation, in this work, we learn a deep representation of irregular time series data by effectively leveraging both missing indicators and time interval. We consider these sources of missingness as a human knowledge such as what to measure and when to measure in different situations, which are indirectly represented in the data. In this context, we regard the representations of missing indicators and time interval as multi-view features of irregularly-sampled observation. Specifically, we propose a multi-view features integration learning for modeling the inter-relations among multi-view observations. This is achieved by using the self-attention mechanism, where the inner product of representations often reflects relationship such as similarity.

}

\subsubsection{Input and Time Embedding}
Given the $D$ measurements at every time step, the first step is to learn the respective input embeddings for observation, missing indicators and time interval. Compared to \cite{vaswani2017attention} that only considered sequence order rather than temporal patterns in positional encoding, in this work, we employ a time embedding as a variant of positional encoding that takes continuous time values as input, and convert them into an encoding vector representation. This approach deals with irregularly-sampled time series by considering the exact time points and their time interval. For time embedding (TE), sine and cosine functions proposed in \cite{vaswani2017attention} are modified as follows:
\begin{align}
    \operatorname{TE}_{(t, 2{ d})} &= \sin{\left(t / l_{max}^{2{ d}/d_{model}}\right)}\\
    \operatorname{TE}_{(t, 2{ d}+1)} &= \cos \left(t/l_{max}^{2{ d}/d_{model}} \right)
\end{align}
where $t$, ${ d}$, $d_{model}$, and $l_{max}$ denote the exact time point, variable index, the dimension of model, and the maximum time length of data, respectively. The time embeddings are added to the learned input embeddings.

\subsubsection{Self-attention}
The basic building block is based on multi-head self-attention (MHA), where a scaled dot-product attention is calculated on a set of queries ($\mathbf{Q}$), keys ($\mathbf{K}$), and values ($\mathbf{V}$) as follows:
\begin{gather}
	\boldsymbol{\alpha}(\mathbf{Q},\mathbf{K},\mathbf{V}) = \sigma\left(\frac{\mathbf{Q} \mathbf{K}^{\top}}{\sqrt{d_k}}\right) \mathbf{V}
\end{gather}
where $\boldsymbol{\alpha}$ denotes the attention function, $\sigma$ is softmax activation function, $d_k$ is the dimension of the key vector.

\begin{table*}[tbp]
\centering
 \caption{Results of the proposed method and the competing methods for mortality prediction ($m\pm \sigma$ from 5-fold cross-validation)}
\renewcommand{\arraystretch}{1}
\setlength{\tabcolsep}{6pt}
\scalebox{0.795}
{
\begin{tabular}{llcccc}
    \toprule
    \multirow{2}{*}{\textbf{Catogory}} & \multirow{2}{*}{\textbf{Method}}  	& \multicolumn{2}{c}{\textbf{MIMIC-III}}  &\multicolumn{2}{c}{\textbf{PhysioNet}} \\
                        												& & AUC   & AUPRC     & AUC               & AUPRC\\\toprule
    \multirow{5}{*}{\textbf{\textit{Recurrence-based Method}}} 	&T-LSTM \cite{tlstm} & {0.8216 $\pm$ 0.0205}    & {0.3475 $\pm$ 0.0430}    & {0.8012 $\pm$ 0.0205}    & {0.3932 $\pm$ 0.0463}\\
    							& GRU-D \cite{GRUD}     & {0.8371 $\pm$ 0.0363}    & {0.3662 $\pm$ 0.0123}        & {0.7990 $\pm$ 0.0223}   & {0.4215 $\pm$ 0.0261} \\
    							& ATTAIN \cite{attain}     & {0.8302 $\pm$ 0.0312}    & {0.3572 $\pm$ 0.0304}        & {0.8093 $\pm$ 0.0318}   & {0.4027 $\pm$ 0.0394} \\
   							& DATA-GRU \cite{tan2020data} & {0.8378 $\pm$ 0.0117}    & {0.3584 $\pm$ 0.0153}    & {0.7982 $\pm$ 0.0160}    & {0.3977 $\pm$ 0.0509}\\
   							& ODE-LSTM \cite{odelstm} & {0.8423 $\pm$ 0.0077}   & {0.3513 $\pm$ 0.0147}    & {0.8065 $\pm$ 0.0395}    & {0.4190 $\pm$ 0.0425}\\\midrule
    
    \multirow{2}{*}{\textbf{\textit{Interpolation-based Method}}} & MGP-RNN \cite{MGPRNN}      & {0.8115 $\pm$ 0.0194}    & {0.3354 $\pm$ 0.0377}        & {0.7970 $\pm$ 0.0244}   & {0.3754 $\pm$ 0.0479} \\
    							& IPNet \cite{InterpNet1} & {0.8226 $\pm$ 0.0261}    & {0.3592 $\pm$ 0.0525}        & {0.7518 $\pm$ 0.0295}   & {0.3063 $\pm$ 0.0177} \\ \midrule
    
   	\multirow{2}{*}{\textbf{\textit{Attention-based Method}}}	& SAnD \cite{AttnDiag}  & {0.8256 $\pm$ 0.0159}    & \textbf{0.3712 $\pm$ 0.0220}        & {0.7911 $\pm$ 0.0373}   & {0.4327 $\pm$ 0.0543} \\    
    
    							& \textbf{Proposed}    & \textbf{0.8534 $\pm$ 0.0071}    & {0.3565 $\pm$ 0.0133}    & \textbf{0.8231 $\pm$ 0.0221}    & \textbf{0.4730 $\pm$ 0.0477}\\ \toprule
    \end{tabular}
    }
    \label{tb:prediction_performance}
\end{table*}

\subsubsection{Multi-view Integration Attention}
Based on the MHA block, we learn the attention representations of multi-view irregular time series including observation, masking indicators, and time interval. { Specifically, each input set ($\mathbf{X}$, $\mathbf{M}$, $\mathbf{\Delta}$) learns its own representation ($\mathbf{H}^{\mathbf{X}}$, $\mathbf{H}^{\mathbf{M}}$, $\mathbf{H}^{\mathbf{\Delta}}$) through self-attention, where each data is feed corresponding to $\mathbf{Q}$, $\mathbf{K}$, and $\mathbf{V}$}.

In this work, we devise a novel multi-view integration attention module (MIAM) that consists of two submodules: (i) an integration module that relies mostly on the self-attention mechanism, and (ii) a position-wise fully connected feed-forward network (FFN) module. Integration module aims to learn complex missing pattern by integrating missing indicators and time interval, and further combine observation and the learned missing pattern in the representation space. { While the most works in the literature \cite{GRUD,lipton2016directly,tlstm,attain,tan2020data} exploited either missing indicators or time interval, and applied the heuristic decaying function for its use, we effectively learn informative missing pattern by using both missing indicators and time interval in the representation space. We argue that learning the underlying representation from missingness itself removes the need to impute values, and does not require specifying any heuristic function.} 

{ In the integration module, there are two integration steps}, \ie, missingness integration and observation-missingness integration. In the missingness integration step, we incorporate the representation of missing indicators with that of time interval by MHA block in Eq. \eqref{eq:missing_attention}, resulting into the representation of missing pattern ($\mathbf{H}^{\mathbf{M}^*}$). 
 \begin{equation}
	\mathbf{H}^{\mathbf{M}^*}=\boldsymbol{\alpha}(\mathbf{H}^{\mathbf{\boldsymbol{\Delta}}}, \mathbf{H}^{\mathbf{M}},\mathbf{H}^{\mathbf{M}}) = \sigma\left(\frac{\mathbf{H}^{\mathbf{\boldsymbol{\Delta}}} {\mathbf{H}^{\mathbf{M}}}^{\top}}{\sqrt{d_k}}\right) \mathbf{H}^{\mathbf{M}} \label{eq:missing_attention} 
\end{equation}
Similarly, in the observation-missingness integration step, the representation of observation is combined with that of missing pattern by computing another MHA block in Eq. \eqref{eq:obs_attention}. 
 \begin{equation}
	\mathbf{H}^{\mathbf{X}^*}=\boldsymbol{\alpha}(\mathbf{H}^{\mathbf{M}^*},\mathbf{H}^{\mathbf{X}},\mathbf{H}^{\mathbf{X}}) = \sigma\left(\frac{\mathbf{H}^{\mathbf{M}^*} {\mathbf{H}^{\mathbf{X}}}^{\top}}{\sqrt{d_k}}\right) \mathbf{H}^{\mathbf{X}} \label{eq:obs_attention}
\end{equation}

This final attention output is the jointly learned deep representations that model the relation between irregular observation data and missing pattern. Then FFN module is applied to each time point identically for modeling the dependency among variables. For more powerful deep representations, the MIAM modules are stacked $L$ times, and the output obtained at the end of the MIAM becomes $\mathbf{H}^{\mathbf{X}}$ again. The final representation $\mathbf{H}^{\mathbf{X}^*}$ is leveraged for the downstream prediction and auxiliary imputation tasks.

\subsection{In-hospital Mortality Prediction}
To predict the probability of in-hospital mortality, given the time series representation $\mathbf{H}^{\mathbf{X}^*}$, we conduct average pooling over timestamps, which results into a final pooled representation $\tilde{\mathbf{h}}$, and apply a multi-layer perceptron (MLP), followed by a sigmoid activation function as follows:
\begin{equation}
    p(y = 1 | \tilde{\mathbf{h}} ) = {\varphi_2}({\varphi_1}(\tilde{\mathbf{h}}\mathbf{W}_{1}+\mathbf{b}_{1})\mathbf{W}_{2}+\mathbf{b}_{2})
\end{equation}
where $\{\mathbf{W}_1,\mathbf{b}_1,\mathbf{W}_2,\mathbf{b}_2\}$ are parameter matrices for classification, and {$\varphi_1$ and $\varphi_2$} are a LeakyReLU activation and sigmoid activation function, respectively.

Furthermore, to handle poor classification performance problem in highly imbalanced data found in healthcare dataset, we employ focal loss {\cite{lin2017focal,9177349}} as the objective function to calculate the classification loss between the target mortality label $y$ and the predicted label $\hat{y}$ or each patient:
\begin{equation}
    \mathcal{L}_{cls} = \sum_{{ n}=1}^{N} -\beta (1- \hat{y}^{({ n})})^{\gamma} \log( \hat{y}^{({ n})})
\end{equation}
where $N$ is the total number of patients, $\gamma$ is a focusing parameter for a minority class, and $\beta$ is a weighting factor to balance the importance between { classes}.

\subsection{Auxiliary Missing Data Imputation}
\label{subsec:imputer}
{ On top of the MIAM module, we also build an attention-based decoder as a missing data imputer that aims to enhance representational power of the inter-relations among multi-view observations for the prediction task. To investigate the masked imputation loss, we randomly masked 10\% of non-missing values and predicted them. From a self-supervised learning perspective, it can be regarded as similar to a masked language modeling task used by BERT \cite{devlin2018bert} that randomly masks some tokens in a text sequence, and then independently recovers the masked tokens to learn language representations. By taking a similar approach, we learn the inter-relations between corrupted values and context, which further contributes to learning the inter-relations among multi-view observations. It should be noted that the proposed method is basically an imputation-free method since imputation only operates in the training phase, not the test phase. Therefore, it has the advantage of reducing model complexity, while not facing the existing imputation-related problems mentioned earlier.

In the attention-based decoder, we further apply an attention block between the final representation of the MIAM module ($\mathbf{H}^{\mathbf{X}^*}$) as query and observation representation ($\mathbf{H}^{\mathbf{X}}$) as key and value, followed by FFN. The output layer maps the output of FFN to the target time sequence by the learned embedding, and results into the imputed data ($\hat{\mathbf{X}}$).

Given another masking vector $\mathbf{M}_{\text{imp}}$ introduced for the purpose of marking the masked values, the imputation loss $\mathcal{L}_{\text{imp}}$ is calculated by \emph{masked mean squared error (MSE)} between the original sample $\bX$ as the ground truth and the imputed sample $\hat{\mathbf{X}}$ for the marked values only by $\bM_{\text{imp}}$ as:
\begin{gather}\tag{24}
	\mathcal{L}_{\text{imp}} = \sum_{{ n}=1}^{N} \frac{\left({\bX}^{({ n})}\odot{\bM}_{\text{imp}}^{({ n})}-\hat{\mathbf{X}}^{({ n})}\odot{\bM}_{\text{imp}}^{({ n})}\right)^2}{N}.
\end{gather}
By adding imputation loss to the objective function, the imputer provides auxiliary information to achieve the optimal prediction result.}

\subsection{Loss Function}
\label{subsec:objective}
{ The composite loss is defined by accumulating the prediction and imputation losses as $\mathcal{L}=\lambda_{\text{imp}}\mathcal{L}_{\text{imp}}+\lambda_{\text{cls}}\mathcal{L}_{\text{cls}}$, where $\lambda_{\text{imp}}$ and $\lambda_{\text{cls}}$ are hyper-parameters that control the ratio between two losses. We optimize all the parameters of our model in an end-to-end manner via the composite loss $\mathcal{L}$.}

\section{Experiment} 
In this section, we evaluated our proposed method for the in-hospital mortality prediction on two publicly available datasets: the Medical Information Mart for Intensive Care III (MIMIC-III), and the PhysioNet 2012 challenge dataset. We reported the average results of the area under the ROC curve (AUC) and the area under the precision–recall curve (AUPRC) from 5-fold cross validation, comparing with other state-of-the-art (SOTA) methods in the literature. In addition, we conducted extensive ablation studies to evaluate the effect of multi-view integration module and our auxiliary imputer.

\subsubsection{Dataset and Preprocessing}
\label{subsec:Dataset}
 We conducted experiments on two datasets: MIMIC-III\footnote{Available at \url{https://mimic.physionet.org/}.} and PhysioNet 2012 challenge\footnote{Available at \url{https://physionet.org/content/challenge-2012/}.} datasets. MIMIC-III consists of medical records of 13,998 patients collected at Beth Israel Deaconess Medical Center. We used 99 different time series measurements for each patient. The selected time series was scarcely observed leading to a missing rate of approximately 90\%. For the in-hospital mortality label, the ratio between 1,181 positive (dead in hospital) and 12,817 negative (alive in hospital) was approximately 1:10.8. The number of irregular time points ranges from 1 to 247 ($m\pm\sigma: 49.29 \pm 35.90$).

PhysioNet 2012 challenge consists of 35 different time-series measurements and 5 general descriptors for 4,000 critical care patients with at least 48 hours of hospital stay. We used 35 time-series variables without the general descriptor of 3,997 patients with at least one observation time. The missing rate is approximately 80.5\%, and the mortality labels are imbalanced at a ratio of approximately 1:6 between 554 positive and 3,443 negative cases. The number of irregular time points ranges from 1 to 202 ($m\pm\sigma: 73.87 \pm 23.06$).

In terms of data preprocessing, because each variable has a different range, all inputs were first Winsorized for removing outliers and then $z$-normalized using the global mean and standard deviation from the entire training set to achieve a zero mean and unit variance in a variable-wise manner.

\subsubsection{Experimental Settings}
\label{subsec:Preprocessing_and_Training}
We trained our models using the Rectified Adam (RAdam) optimizer \cite{liu2019variance} with an initial learning rate of $0.005$ and a multiplicative decay of $0.2$ every $10$ epochs for $60$ epochs using minibatches of $64$ samples. We chose the final optimal model based on the performance of the validation set.

For MIMIC-III dataset, the model achieved its best performance at two-layered MHA with 8 heads of $d_k=d_v=128$, and $d_{FFN}=128$. For PhysioNet 2012 challenge dataset, we observed the best performance at two-layered MHA with 8 heads of $d_k=d_v=64$, and $d_{FFN}=128$. The $\beta$ and $\gamma$ in focal loss were chosen 7 and 0.15, respectively, and $\lambda_{\text{imp}}$ and $\lambda_{\text{cls}}$ in the composite loss were 0.1 and 7, respectively.

We validated the efficacy of our proposed method by comparing it with SOTA methods categorized as recurrence-based method, interpolation-based method, and attention-based method as shown in Table \ref{tb:prediction_performance}.

\section{Results}
\label{sec:Results}
\subsubsection{Performance Comparison}
\label{subsec:performance_comparison}
Table \ref{tb:prediction_performance} compares the results of our proposed method with those of the competing methods for mortality prediction on MIMIC-III and PhysioNet 2012 challenge datasets. For both datasets, our model achieved the best classification performance. Overall performance of recurrence-based methods and attention-based methods were similar, and interpolation methods showed relatively lower performance than the others. Among the recurrence-based methods, ODE-LSTM \cite{odelstm} demonstrated a competitive performance, which is a continuous time LSTM-based model that takes irregularly-sampled time series as input and simultaneously handles time intervals. It suggests the need for directly modeling irregularly sampled time series and a more sophisticated use of missing indicators and time interval. These experimental results validated the efficacy of our proposed method that learns the multi-view representation of irregular time series data and their deep integration with the self-attention mechanism, and further builds the auxiliary missing data imputer, showing its superior performance in the downstream task.

\begin{table}[tb]
	\setlength{\tabcolsep}{6pt}
	\renewcommand*{\arraystretch}{1}
	\caption{Results of ablation studies in terms of the multi-view integration ($m\pm \sigma$ from 5-fold cross-validation). Among the methods, the bold text indicates the proposed approach.}
	 \begin{center}
	\scalebox{0.76}{
	\begin{tabular}{llcccc}
	\toprule
	\textbf{Dataset} 				& \textbf{Method} 				& \textbf{AUC}   			& \textbf{AUPRC}     \\\toprule
	\multirow{4}{*}{\textbf{MIMIC-III}}	& Single ($\mathbf{X})$         & {0.7943 $\pm$ 0.0048}    	& {0.3044 $\pm$ 0.0091}  \\ 
								& Double ($\mathbf{X} + \mathbf{M}$) & {0.8273 $\pm$ 0.0058}    & {0.3385 $\pm$ 0.0170}     \\
								& Double ($\mathbf{X} + \mathbf{\Delta}$)    & {0.8458 $\pm$ 0.0051}    & {0.3577 $\pm$ 0.0155}   \\ 
								& \textbf{Triple ($\mathbf{X} + \mathbf{M} + \mathbf{\Delta}$)}   & \textbf{0.8534 $\pm$ 0.0071}    & \textbf{0.3565 $\pm$ 0.0133}  \\ \midrule                         
								          		
	\multirow{4}{*}{\textbf{PhysioNet}}	& Single ($\mathbf{X})$		& {0.7816 $\pm$ 0.0381}   & {0.4195 $\pm$ 0.0345} \\ 
								& Double ($\mathbf{X} + \mathbf{M}$)  & {0.8131 $\pm$ 0.0292}   & {0.4475 $\pm$ 0.0452} \\
								& Double ($\mathbf{X} + \mathbf{\Delta}$)    & {0.8142 $\pm$ 0.0175}    & {0.4501 $\pm$ 0.0225}\\ 
								& \textbf{Triple ($\mathbf{X} + \mathbf{M} + \mathbf{\Delta}$)}    & \textbf{0.8231 $\pm$ 0.0221}    & \textbf{0.4730 $\pm$ 0.0477} \\ 	\toprule	
\end{tabular}
}
\label{tb:ablation1}
\end{center}
\end{table}

\begin{table}[b]
	\setlength{\tabcolsep}{12pt}
	\renewcommand*{\arraystretch}{1}
	\caption{Results of ablation studies related to the attention-based imputer ($m\pm \sigma$ from 5-fold cross-validation)}
	 \begin{center}
	\scalebox{0.74}{
	\begin{tabular}{llcccc}
	\toprule
	\textbf{Dataset} 				& \textbf{Method} 							& \textbf{AUC}   			& \textbf{AUPRC}     \\\toprule
	\multirow{2}{*}{\textbf{MIMIC-III}}	& w/o Imputer   						& {0.8373 $\pm$ 0.0077}  		& {0.3487 $\pm$ 0.0147}    \\
								& \textbf{w/ Imputer}   						& {0.8534 $\pm$ 0.0071}    	& {0.3565 $\pm$ 0.0133}  \\ \midrule                                 		
	\multirow{2}{*}{\textbf{PhysioNet}}	& w/o Imputer     						& {0.8137 $\pm$ 0.0232}    	& {0.4515 $\pm$ 0.0416} \\
								& \textbf{w/ Imputer}    						& {0.8231 $\pm$ 0.0221}    	& {0.4730 $\pm$ 0.0477}\\ \toprule 
\end{tabular}
}
\label{tb:ablation2}
\end{center}
\end{table}

\subsubsection{Ablation Study}
\label{subsec:Ablation_Study}
In this study, we conducted extensive ablation studies to investigate the influence of different experimental design options in terms of multi-view integration module and auxiliary imputer in our method. Table \ref{tb:ablation1} shows the results of investigating the effectiveness of multi-view integration through four scenarios: single view ($\mathbf{X}$), double views ($\mathbf{X} + \mathbf{M}$, $\mathbf{X} + \mathbf{\Delta}$), and triple views ($\mathbf{X} + \mathbf{M} + \mathbf{\Delta}$). It can be seen that integrating observation and sources of missingness was better than using observation only, and the result of integration for triple views was the highest in both MIMIC-III and PhysioNet 2012 datasets, which shows the effectiveness of incorporating observation and sources of missingness. In the double view cases, the scenario of ($\mathbf{X} + \mathbf{\Delta}$) achieved slightly better performance than that of ($\mathbf{X} + \mathbf{M}$) in both datasets.

In addition, we explored the effect of the attention-based imputer, as shown in Table \ref{tb:ablation2}. The results showed that the proposed model built with the attention-based imputer achieved higher performance than w/o imputer in both datasets, indicating that our auxiliary imputer helps enhance representational power of the inter-relations among multi-view observations for the prediction task.

Furthermore, we examined the necessity of using explicit imputation data in the prediction task in Table \ref{tb:ablation3}. In this regard, we compared our proposed approach with the case of using the imputed data by BRITS imputer \cite{cao2018brits} as the SOTA imputation method and our auxiliary imputer. In these cases, the imputation (dotted line in Figure \ref{fig:framework}) is also conducted in the test phase. The result of using the imputed data by our auxiliary imputer (Auxiliary Imputation+MIAM) is comparable to that of the proposed approach (No Imputation+MIAM) and that of (BRITS Imputation+MIAM).

These results indicate that using explicit imputation data contributes to the improvement of prediction task performance to some extent rather than without imputation, but its effect is insignificant. Thus, our proposed approach of using the auxiliary imputer in the training phase only suffices to deal with modeling irregular time series data.

\begin{table}[tb]
	\setlength{\tabcolsep}{5pt}
	\renewcommand*{\arraystretch}{1}
	\caption{Results of ablation studies to investigate the effect of using the imputed data by BRITS imputer and Auxiliary imputer ($m\pm \sigma$ from 5-fold cross-validation). In these cases, the imputation (dotted line in Figure \ref{fig:framework}) is also conducted in the test phase. Among the methods, the bold text indicates the proposed approach.}
	 \begin{center}
	\scalebox{0.715}{
	\begin{tabular}{llcccc}
	\toprule
	\textbf{Dataset} 				& \textbf{Method} 							& \textbf{AUC}   			& \textbf{AUPRC}     \\\toprule
	\multirow{3}{*}{\textbf{MIMIC-III}}	& BRITS Imputation+MIAM	& {0.8598 $\pm$ 0.0082}		& {0.3611 $\pm$ 0.0217} \\ 
								& Auxiliary Imputation+MIAM					& {0.8562 $\pm$ 0.0117}		& {0.3580 $\pm$ 0.0153}\\      
								& \textbf{No Imputation+MIAM}   						& {0.8534 $\pm$ 0.0071}    	& {0.3565 $\pm$ 0.0133}  \\ \midrule                                 		
	\multirow{3}{*}{\textbf{PhysioNet}}	& BRITS Imputation+MIAM	& {0.8305 $\pm$ 0.0156}		& {0.4765 $\pm$ 0.0360}\\ 
								& Auxiliary Imputation+MIAM					& {0.8253 $\pm$ 0.0235}		& {0.4742 $\pm$ 0.0482}\\
								& \textbf{No Imputation+MIAM}    						& {0.8231 $\pm$ 0.0221}    	& {0.4730 $\pm$ 0.0477}\\ \toprule 
\end{tabular}
}
\label{tb:ablation3}
\end{center}
\end{table}
\section{Conclusion}

In this work, we proposed a method to directly learn the integrated representations of multi-view features from irregular multivariate time series data by using the self-attention mechanism without imputation. Specifically, we devised a novel multi-integration attention module (MIAM) to extract complex missing pattern by integrating missing indicators and time interval, and further combine observation and missing pattern in the representation space through a series of self-attention blocks. In addition, we built an attention-based decoder as a missing value imputer that helps empower the representation learning of the inter-relations among multi-view observations for prediction task, which operates at the training phase only. We validated the effectiveness of our method over the public MIMIC-III and PhysioNet challenge 2012 datasets by comparing with and outperforming the state-of-the-art methods for in-hospital mortality prediction.

\section{Acknowledgments}
This work was supported by Institute of Information $\&$ communications Technology Planning $\&$ Evaluation (IITP) grant funded by the Korea government(MSIT) (No. 2019-0-00079, Artificial Intelligence Graduate School Program(Korea University))

\bibliographystyle{named}
\bibliography{ijcai21}

\begin{thebibliography}{}

\bibitem[\protect\citeauthoryear{Baytas \bgroup \em et al.\egroup
  }{2017}]{tlstm}
Inci~M Baytas, Cao Xiao, Xi~Zhang, Fei Wang, Anil~K Jain, and Jiayu Zhou.
\newblock Patient subtyping via time-aware {LSTM} networks.
\newblock In {\em SIGKDD}, pages 65--74, 2017.

\bibitem[\protect\citeauthoryear{Cao \bgroup \em et al.\egroup
  }{2018}]{cao2018brits}
Wei Cao, Dong Wang, Jian Li, Hao Zhou, Lei Li, and Yitan Li.
\newblock {BRITS}: Bidirectional recurrent imputation for time series.
\newblock {\em NeurIPS}, 31:6775--6785, 2018.

\bibitem[\protect\citeauthoryear{Che \bgroup \em et al.\egroup }{2018}]{GRUD}
Zhengping Che, Sanjay Purushotham, Kyunghyun Cho, David Sontag, and Yan Liu.
\newblock Recurrent neural networks for multivariate time series with missing
  values.
\newblock {\em Scientific Reports}, 8(1):1--12, 2018.

\bibitem[\protect\citeauthoryear{Choi \bgroup \em et al.\egroup
  }{2016}]{retain}
Edward Choi, Mohammad~Taha Bahadori, Jimeng Sun, Joshua Kulas, Andy Schuetz,
  and Walter Stewart.
\newblock {RETAIN}: An interpretable predictive model for healthcare using
  reverse time attention mechanism.
\newblock In {\em NIPS}, pages 3504--3512, 2016.

\bibitem[\protect\citeauthoryear{Choi \bgroup \em et al.\egroup
  }{2017}]{choi2017gram}
Edward Choi, Mohammad~Taha Bahadori, Le~Song, Walter~F Stewart, and Jimeng Sun.
\newblock {GRAM}: graph-based attention model for healthcare representation
  learning.
\newblock In {\em SIGKDD}, pages 787--795, 2017.

\bibitem[\protect\citeauthoryear{Devlin \bgroup \em et al.\egroup
  }{2018}]{devlin2018bert}
Jacob Devlin, Ming-Wei Chang, Kenton Lee, and Kristina Toutanova.
\newblock {BERT}: Pre-training of deep bidirectional transformers for language
  understanding.
\newblock {\em arXiv preprint arXiv:1810.04805}, 2018.

\bibitem[\protect\citeauthoryear{Fortuin \bgroup \em et al.\egroup
  }{2020}]{fortuin2020gp}
Vincent Fortuin, Dmitry Baranchuk, Gunnar R{\"a}tsch, and Stephan Mandt.
\newblock {GP-VAE}: Deep probabilistic time series imputation.
\newblock In {\em AISTATS}, pages 1651--1661, 2020.

\bibitem[\protect\citeauthoryear{Futoma \bgroup \em et al.\egroup
  }{2017}]{MGPRNN}
Joseph Futoma, Sanjay Hariharan, and Katherine Heller.
\newblock Learning to detect sepsis with a multitask {G}aussian process {RNN}
  classifier.
\newblock In {\em ICML}, pages 1174--1182, 2017.

\bibitem[\protect\citeauthoryear{Horn \bgroup \em et al.\egroup
  }{2020}]{horn2020set}
Max Horn, Michael Moor, Christian Bock, Bastian Rieck, and Karsten Borgwardt.
\newblock Set functions for time series.
\newblock In {\em ICML}, pages 4353--4363, 2020.

\bibitem[\protect\citeauthoryear{Jun \bgroup \em et al.\egroup
  }{2019}]{jun2019stochastic}
Eunji Jun, Ahmad~Wisnu Mulyadi, and Heung-Il Suk.
\newblock Stochastic imputation and uncertainty-aware attention to {EHR} for
  mortality prediction.
\newblock In {\em IJCNN}, pages 1--7, 2019.

\bibitem[\protect\citeauthoryear{{Jun} \bgroup \em et al.\egroup
  }{2020}]{9177349}
E.~{Jun}, A.~W. {Mulyadi}, J.~{Choi}, and H.~I. {Suk}.
\newblock Uncertainty-gated stochastic sequential model for ehr mortality
  prediction.
\newblock {\em IEEE Transactions on Neural Networks and Learning Systems},
  pages 1--11, 2020.

\bibitem[\protect\citeauthoryear{Lechner and Hasani}{2020}]{odelstm}
Mathias Lechner and Ramin Hasani.
\newblock Learning long-term dependencies in irregularly-sampled time series.
\newblock In {\em NeurIPS}, 2020.

\bibitem[\protect\citeauthoryear{Lin \bgroup \em et al.\egroup
  }{2017}]{lin2017focal}
Tsung-Yi Lin, Priya Goyal, Ross Girshick, Kaiming He, and Piotr Doll{\'a}r.
\newblock Focal loss for dense object detection.
\newblock In {\em ICCV}, pages 2980--2988, 2017.

\bibitem[\protect\citeauthoryear{Lipton \bgroup \em et al.\egroup
  }{2016}]{lipton2016directly}
Zachary~C Lipton, David Kale, and Randall Wetzel.
\newblock Directly modeling missing data in sequences with {RNN}s: Improved
  classification of clinical time series.
\newblock In {\em MLHC}, pages 253--270, 2016.

\bibitem[\protect\citeauthoryear{Liu \bgroup \em et al.\egroup
  }{2019}]{liu2019variance}
Liyuan Liu, Haoming Jiang, Pengcheng He, Weizhu Chen, Xiaodong Liu, Jianfeng
  Gao, and Jiawei Han.
\newblock On the variance of the adaptive learning rate and beyond.
\newblock {\em arXiv preprint arXiv:1908.03265}, 2019.

\bibitem[\protect\citeauthoryear{Ma \bgroup \em et al.\egroup
  }{2018}]{ma2018kame}
Fenglong Ma, Quanzeng You, Houping Xiao, Radha Chitta, Jing Zhou, and Jing Gao.
\newblock {KAME}: Knowledge-based attention model for diagnosis prediction in
  healthcare.
\newblock In {\em CIKM}, pages 743--752, 2018.

\bibitem[\protect\citeauthoryear{{Nguyen} \bgroup \em et al.\egroup
  }{2017}]{Deepr}
P.~{Nguyen}, T.~{Tran}, N.~{Wickramasinghe}, and S.~{Venkatesh}.
\newblock $\mathtt{Deepr}$: A convolutional net for medical records.
\newblock {\em IEEE Journal of Biomedical and Health Informatics},
  21(1):22--30, 2017.

\bibitem[\protect\citeauthoryear{Shukla and Marlin}{2019}]{InterpNet1}
Satya~Narayan Shukla and Benjamin~M Marlin.
\newblock Interpolation-prediction networks for irregularly sampled time
  series.
\newblock {\em arXiv preprint arXiv:1909.07782}, 2019.

\bibitem[\protect\citeauthoryear{Song \bgroup \em et al.\egroup
  }{2018}]{AttnDiag}
Huan Song, Deepta Rajan, Jayaraman~J Thiagarajan, and Andreas Spanias.
\newblock Attend and diagnose: Clinical time series analysis using attention
  models.
\newblock In {\em AAAI}, 2018.

\bibitem[\protect\citeauthoryear{Suo \bgroup \em et al.\egroup
  }{2017}]{timefusionCNN}
Qiuling Suo, Fenglong Ma, Ye~Yuan, Mengdi Huai, Weida Zhong, Aidong Zhang, and
  Jing Gao.
\newblock Personalized disease prediction using a {CNN}-based similarity
  learning method.
\newblock In {\em IEEE BIBM}, pages 811--816, 2017.

\bibitem[\protect\citeauthoryear{Tan \bgroup \em et al.\egroup
  }{2020}]{tan2020data}
Qingxiong Tan, Mang Ye, Baoyao Yang, Siqi Liu, Andy~Jinhua Ma, Terry Cheuk-Fung
  Yip, Grace Lai-Hung Wong, and PongChi Yuen.
\newblock {DATA}-{GRU}: Dual-attention time-aware gated recurrent unit for
  irregular multivariate time series.
\newblock In {\em AAAI}, volume~34, pages 930--937, 2020.

\bibitem[\protect\citeauthoryear{Vaswani \bgroup \em et al.\egroup
  }{2017}]{vaswani2017attention}
Ashish Vaswani, Noam Shazeer, Niki Parmar, Jakob Uszkoreit, Llion Jones,
  Aidan~N Gomez, {\L}ukasz Kaiser, and Illia Polosukhin.
\newblock Attention is all you need.
\newblock In {\em NIPS}, pages 5998--6008, 2017.

\bibitem[\protect\citeauthoryear{Yoon \bgroup \em et al.\egroup
  }{2018}]{yoon2018gain}
Jinsung Yoon, James Jordon, and Mihaela Schaar.
\newblock {GAIN}: Missing data imputation using generative adversarial nets.
\newblock In {\em ICML}, pages 5689--5698, 2018.

\bibitem[\protect\citeauthoryear{Zaheer \bgroup \em et al.\egroup
  }{2017}]{zaheer2017deep}
Manzil Zaheer, Satwik Kottur, Siamak Ravanbakhsh, Barnabas Poczos, Russ~R
  Salakhutdinov, and Alexander~J Smola.
\newblock Deep sets.
\newblock In {\em NIPS}, page 3391–3401, 2017.

\bibitem[\protect\citeauthoryear{Zhang}{2019}]{attain}
Yuan Zhang.
\newblock {ATTAIN}: Attention-based time-aware {LSTM} networks for disease
  progression modeling.
\newblock In {\em IJCAI}, 2019.

\end{thebibliography}

\end{document}